\documentclass[11pt]{article}

\usepackage[preprint]{acl}

\usepackage{times}
\usepackage{latexsym}

\usepackage[T1]{fontenc}
\usepackage[utf8]{inputenc}

\usepackage{microtype}

\usepackage{inconsolata}

\usepackage{tcolorbox} % for tcolorbox
\tcbuselibrary{skins} % for rounded corners

\usepackage{graphicx}
\usepackage{amsmath}
\usepackage{xspace}

\usepackage{amsfonts}
\usepackage{booktabs}
\usepackage{bm}
\newcommand{\rfc}{\mbox{\textsc{ReFoRCE}}}
\newcommand{\dso}{\mbox{\textsc{DeepSeek-OCR}}}
\newcommand{\pix}{\mbox{\textsc{Pix2Struct}}}
\newcommand{\don}{\mbox{\textsc{Donut}}}
\newcommand{\soe}{\mbox{\textsc{OptiSQL (FrozenEnc)}}}
\newcommand{\sof}{\mbox{\textsc{OptiSQL (FullFT)}}}

\definecolor{deeporange}{RGB}{255, 140, 0}
\definecolor{ddgreen}{RGB}{108, 138, 90}
\usepackage{listings}
\usepackage[ruled,vlined]{algorithm2e}

\title{OptiSQL: Executable SQL Generation from Optical Tokens}

\author{
 \textbf{Sifan~Li\textsuperscript{\dag\ddag}}
  \hspace{.8em}
 \textbf{Hongkai~Chen\textsuperscript{\ddag}}
  \hspace{.8em}
 \textbf{Yujun~Cai\textsuperscript{\S}}
 \hspace{.8em}
 \textbf{Liyang~Chen\textsuperscript{\ddag$\flat$}}
  \\
 \textbf{Qingwen~Ye\textsuperscript{\ddag}}
  \hspace{.8em}
 \textbf{Yiwei~Wang\textsuperscript{\dag}}
\\
\\
 \textsuperscript{\dag}University of California, Merced
 \hspace{.8em}
 \textsuperscript{\ddag}vivo Mobile Communication Co., Ltd.
 \\
 \textsuperscript{\S}University of Queensland
 \hspace{.8em}
 \textsuperscript{$\flat$}University of California, Los Angeles
\\
 \texttt{\href{mailto:sflijohn@foxmail.com}{\textcolor{black}{sflijohn@foxmail.com}}}
 \\
\texttt{\href{https://github.com/johnnyZeppelin/OptiSQL}{\textcolor{magenta}{https://github.com/johnnyZeppelin/OptiSQL}}}
}

\begin{document}
\maketitle
\begin{abstract}
Executable SQL generation is typically studied in text-to-SQL settings, where tables are provided as fully linearized textual schemas and contents. While effective, this formulation assumes access to structured text and incurs substantial token overhead, which is misaligned with many real-world scenarios where tables appear as visual artifacts in documents or webpages. We investigate whether compact optical representations can serve as an efficient interface for executable semantic parsing. We present OptiSQL, a vision-driven framework that generates executable SQL directly from table images and natural language questions using compact optical tokens. OptiSQL leverages an OCR-oriented visual encoder to compress table structure and content into a small set of optical tokens and fine-tunes a pretrained decoder for SQL generation while freezing the encoder to isolate representation sufficiency. Experiments on a visualized version of Spider 2.0-Snow show that OptiSQL retains strong execution accuracy while reducing table input tokens by an order of magnitude. Robustness analyses further demonstrate that optical tokens preserve essential structural information under visual perturbations.
\end{abstract}

\section{Introduction}
\label{sec:intro}

\begin{figure*}[t]
  \includegraphics[width=\linewidth]{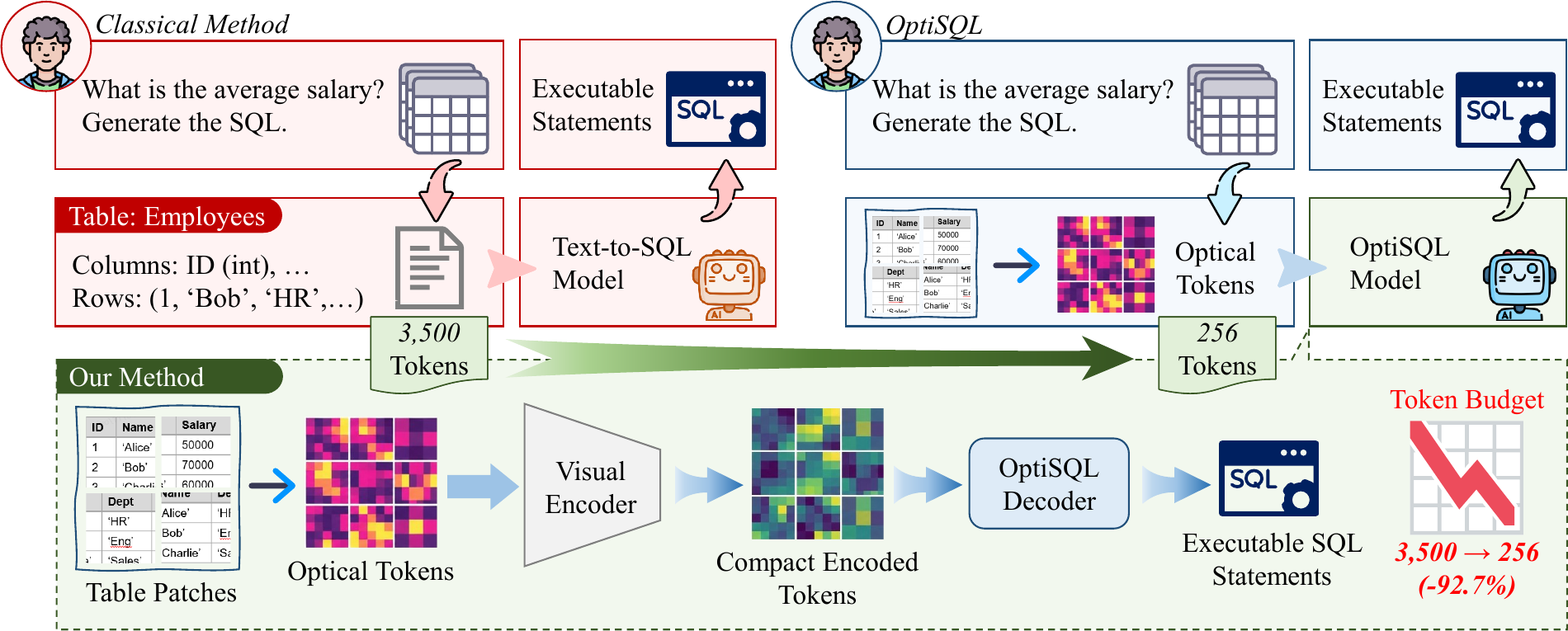}
  \caption{OptiSQL generates executable SQL directly from table images using compact optical tokens. A frozen OCR-oriented visual encoder converts a table image into a fixed-size sequence of optical tokens, which are combined with a natural language question and processed by a trainable decoder. By operating under an explicit token budget, OptiSQL enables an efficient and robust alternative to text-based table encodings.}
  \label{fig:overview}
\end{figure*}

Executable SQL generation has been predominantly studied under the text-to-SQL paradigm, where database schemas and table contents are provided as structured textual inputs \cite{yu2018spider,spider2_arxiv}. While effective in controlled settings, this formulation relies on a strong assumption: that clean, machine-readable table text is readily available. In many real-world scenarios, however, tabular data is encountered primarily in visual form, such as scanned documents, PDFs, or web pages, where structured textual schemas are not directly accessible. Converting such inputs into text typically requires multi-stage preprocessing and manual normalization, making text-based pipelines brittle in practice \cite{donut,p2s}.

This mismatch motivates treating tables as visual objects rather than inherently textual ones. Table images naturally encode row-column structure, alignment, and grouping, which are expensive to represent through token-by-token linearization. Recent OCR-oriented vision-language models have shown that document images can be compressed into compact sequences of discrete visual representations while preserving fine-grained semantic and structural information \cite{deepseekocr}. In particular, DeepSeek-OCR demonstrates that high-resolution document images can be mapped into a small number of optical tokens suitable for downstream language modeling \cite{deepseekocr}. Importantly, these optical tokens capture more than raw text, retaining layout and relational cues critical for table understanding \cite{deepseekocr}.

In this work, we leverage optical tokens as an explicit interface for executable semantic parsing. Rather than reconstructing text or schemas, we ask whether a compact visual abstraction of a table is sufficient to support SQL generation under strict token budgets. We introduce \textbf{OptiSQL}, a vision-driven framework that generates executable SQL queries directly from table images and natural language questions. OptiSQL uses a pretrained OCR-oriented visual encoder to convert a table image into a fixed-length sequence of optical tokens, and fine-tunes a pretrained autoregressive decoder to generate SQL. The visual encoder is frozen by default, isolating the representational sufficiency of optical tokens and avoiding confounding effects from encoder adaptation.

We evaluate OptiSQL on a visualized version of Spider~2.0 \cite{spider2_arxiv}, using execution accuracy as the primary metric. Through controlled analyses of token budgets and visual perturbations, we show that OptiSQL substantially reduces input length while maintaining strong executable accuracy, and that performance degrades predictably as visual information is removed or corrupted. These results indicate that optical tokens provide a practical and analyzable interface for efficiency-oriented executable semantic parsing, especially under long-context cost constraints \cite{liu2023lostmiddle}.
In summary, this work makes the following contributions:
\begin{itemize}
    \item We propose OptiSQL, a vision-driven framework that replaces explicit textual table encodings with compact optical tokens for executable SQL generation.
    \item We introduce a controlled evaluation setting that isolates the sufficiency of optical tokens by freezing the visual encoder and adapting only the decoder.
    \item We demonstrate favorable trade-offs between input efficiency, robustness, and execution accuracy on the semantic parsing task.
\end{itemize}

\section{Related Work}
\label{sec:related}

\subsection{Text-to-SQL Semantic Parsing}
\label{related:t2s}
Text-to-SQL semantic parsing generates executable SQL queries from natural language questions given textual representations of database schemas and contents \cite{yu2018spider, shi-etal-2025-gen, zhou2024dbgpthubopenbenchmarkingtexttosql, spider2_arxiv}. Prior work has improved accuracy through schema linking, constrained decoding, and execution-guided learning \cite{guo2019complextexttosqlcrossdomaindatabase,wang2025querylevelcomparisonfinegrainedreinforcement,liu2023lostmiddle,cai2025text2sqlflowrobustsqlawaredata}. Recent agent-based systems further leverage iterative refinement and execution feedback, with ReFoRCE representing a strong state-of-the-art approach \cite{reforce}. 

These methods assume clean and complete textual access to schemas and table contents, removing the representational bottleneck that arises in visually grounded settings. We therefore treat text-to-SQL systems as \textbf{upper-bound baselines} rather than direct competitors, as they operate under a fundamentally different input assumption.

\subsection{OCR-Centric Pipelines}
\label{related:ocr}
Querying tables embedded in document images is often addressed via OCR-centric pipelines that reconstruct text and table structure before applying text-based parsers \cite{li-etal-2020-tablebank,doc2table,sheng2024pdftableunifiedtoolkitdeep}. While modular, such pipelines are sensitive to recognition errors and typically discard spatial and layout information during text linearization, leading to error propagation and high input costs \cite{guan-etal-2025-prep}. We include an OCR pipeline as a reference baseline to highlight the trade-offs between explicit text reconstruction and efficient executable reasoning.

\subsection{Optical Tokenization for Document Understanding}
\label{related:vlm}
Vision-language models such as Pix2Struct and Donut map document images directly to structured outputs and have been applied to table-related tasks \cite{p2s,donut}. More recently, OCR-oriented visual encoders have been proposed to compress document images into compact sequences of optical tokens that preserve semantic and structural information under strict length budgets \cite{deepseekocr}. DeepSeek-OCR exemplifies this direction, demonstrating that high-resolution document images can be represented by a small number of informative visual tokens.

Our work builds on optical tokenization as a representation interface but targets a different question: whether compact optical tokens are sufficient for \textbf{executable semantic parsing}. Unlike prior work, we explicitly freeze the visual encoder by default and adapt only the decoder, enabling a controlled study of representation sufficiency and efficiency-accuracy trade-offs in SQL generation.

\section{Task Definition}
\label{sec:task}

We study \textbf{executable SQL generation from table images}. Given a table image $I$ and a natural language question $q$, the goal is to generate an executable SQL query $s$ whose execution on the underlying database returns the correct answer.

\subsection{Problem Formulation}
\label{task:formulation}
Unlike standard text-to-SQL settings, the model is \emph{not} given ground-truth textual schemas, column metadata, or table contents as structured text. All table-related information must be inferred from the visual input. We assume the table image captures the schema cues and cell contents needed to answer the question, and the model does not access database metadata or schema annotations at inference time.

\subsection{Efficiency-Oriented Objective}
\label{task:eff}
Our central objective is \textbf{input efficiency}. Let $|V|$ be the number of optical tokens representing $I$ produced by an OCR-oriented visual encoder, and let $|T|$ be the number of tokens required to represent the same table via textual linearization. We study whether accurate SQL generation remains feasible when $|V|\!\ll\!|T|$, and how executable performance varies as we explicitly control the token budget.

\subsection{Evaluation}
\label{task:eval}
We primarily evaluate with execution accuracy (EXAcc), and additionally report canonical exact match (EX-Can) and efficiency metrics such as optical tokens per query and token savings ratio (TSR); metric definitions are provided in Section~\ref{sec:exp}.

\section{Model Architecture}
\label{sec:model}

OptiSQL is a vision-driven semantic parsing framework that generates executable SQL queries from table images under explicit input efficiency constraints. As shown in Figure~\ref{fig:overview}, the model consists of three components: (i) a pretrained OCR-oriented visual encoder that converts a table image into a compact sequence of optical tokens, (ii) a unified token interface that combines optical tokens with the natural language question, and (iii) a pretrained decoder adapted for SQL generation.

A key design principle of OptiSQL is to treat optical tokenization as a fixed representation interface. By default, the visual encoder is frozen and only the decoder is fine-tuned for SQL generation. This setting, denoted as \textsc{FrozenEnc}, isolates the representational capacity of compact optical tokens and enables a controlled study of whether such representations are sufficient for executable semantic parsing. We additionally consider a \textsc{FullFT} variant as an ablation to study accuracy and robustness trade-offs introduced by encoder adaptation. Unless otherwise specified, OptiSQL refers to the \textsc{FrozenEnc} setting throughout the paper.

\subsection{Optical Tokenization via a Visual Encoder}
\label{model:vis}

Given a table image $I$, OptiSQL employs a pretrained OCR-oriented visual encoder $E$ to produce a sequence of optical tokens:
\begin{equation}
    V = E(I) = (v_1, v_2, \dots, v_n),
\end{equation}
where each token encodes localized visual, textual, and structural information extracted from the image. The encoder compresses the two-dimensional table layout into a one-dimensional token sequence while preserving regularities such as row-column alignment, cell grouping information, and header-content associations.

Importantly, OptiSQL does not perform explicit text reconstruction or OCR decoding during inference. The optical tokens are treated as a latent interface that implicitly represents table content and structure. By freezing the visual encoder, we prevent task-specific adaptation and directly evaluate whether the pretrained optical tokens retain sufficient information for executable reasoning.

The number of optical tokens $n$ is controlled by the encoder configuration, enabling systematic exploration of different token budgets. This explicit control is central to our analysis of efficiency-accuracy trade-offs.

\subsection{Unified Input Representation}
\label{model:input}

The decoder input is constructed by concatenating the optical tokens with the tokenized natural language question:
\begin{equation}
    X = [V; Q],
\end{equation}
where $Q$ denotes the question tokens. Optical and textual tokens are projected into a shared embedding space before being processed by the decoder.

No textual schema annotations, column names, or table contents are provided. All table-related information must be inferred from the optical tokens. This design enforces a strict representational bottleneck and distinguishes OptiSQL from text-to-SQL systems and OCR-based pipelines.

\subsection{Autoregressive SQL Decoder}
\label{model:dec}

OptiSQL uses a pretrained decoder $D$ to generate SQL queries conditioned on the combined input sequence $X$. At each decoding step $t$, the model predicts the next SQL token $s_t$ according to:
\begin{equation}
    p(s_t \mid s_{<t}, X).
\end{equation}

The decoder is responsible for producing syntactically valid and executable SQL queries. SQL generation terminates when an end-of-sequence token is produced. While syntax constraints may optionally be applied, the core model relies on learned representations rather than hand-crafted rules.

Under the default \textsc{FrozenEnc} setting, only the decoder parameters are updated during training. This design focuses learning on mapping compact optical representations to executable programs, without altering the visual tokenization itself.

\subsection{Training Variants}
\label{model:training-variants}

We consider two training variants to disentangle representation sufficiency from encoder adaptation effects.

\paragraph{\textsc{FrozenEnc}.}
In the default setting, the visual encoder is frozen and only the decoder is fine-tuned. This variant serves as the primary configuration for evaluating input efficiency, token budget scaling, visual grounding, and robustness.

\paragraph{\textsc{FullFT}.}
As an ablation, we fine-tune both the visual encoder and the decoder. In this setting, we additionally apply robustness-oriented training-time table rendering augmentations. \textsc{FullFT} is used solely to study the trade-offs between clean accuracy and generalization, and is not the default configuration.

\section{Experimental Setup}
\label{sec:exp}

This section describes the datasets, baselines, evaluation metrics, and experimental protocols. All experiments are designed to address a single question: are compact optical tokens sufficient to support executable SQL generation under strict input efficiency constraints? Unless otherwise specified, all OptiSQL results correspond to the \textsc{FrozenEnc} setting, which freezes the encoder.

\subsection{Dataset and Task Construction}
\label{exp:dataset}

We evaluate OptiSQL on a visualized version of Spider~2.0-Snow~\cite{spider2_arxiv}, a realistic enterprise-oriented text-to-SQL benchmark with complex SQL queries and diverse schemas. For each example, the original table is rendered into a table image, while the natural language question and ground-truth SQL query remain unchanged. No textual schema or table content is provided to the model.

\subsection{Baselines}
\label{exp:baseline}

We compare our proposed OptiSQL against representative baselines that reflect different assumptions about table access.

\paragraph{Text-based Text-to-SQL.}
We include ReFoRCE~\cite{reforce}, a state-of-the-art text-to-SQL system operating on ground-truth textual schemas and contents. This setting removes the representational bottleneck entirely and therefore serves as an \textbf{upper-bound baseline}, albeit with very high token cost.

\paragraph{OCR-based Text-to-SQL Pipeline.}
We construct a two-stage pipeline that first applies OCR to recover table text and structure, followed by a text-based SQL parser. This baseline reflects a common document understanding workflow and highlights the effects of OCR errors, schema linearization, and long textual inputs.

\paragraph{Vision-Language Models.}
We evaluate general-purpose vision-language models (VLMs), including Pix2Struct~\cite{p2s} and Donut~\cite{donut}, which directly generate SQL from images without explicit OCR pipelines.

\subsection{Evaluation Metrics}
\label{exp:metrics}

We evaluate models along four dimensions: executable correctness, syntactic fidelity, input efficiency, and robustness.

\paragraph{Execution Accuracy (EXAcc).}
Execution accuracy measures whether the generated SQL query produces the correct result when executed against the database:
\begin{equation}
\text{EXAcc}\!=\!\frac{1}{N}\sum_{i=1}^{N}\mathbf{1}\!\left[\text{EX}(s_i,\mathcal{D}_i)\!=\!\text{EX}(s_i^\ast,\mathcal{D}_i)\right].
\end{equation}
Execution-based evaluation is preferred as semantically equivalent SQL queries may differ in surface form.

\begin{figure}[t]
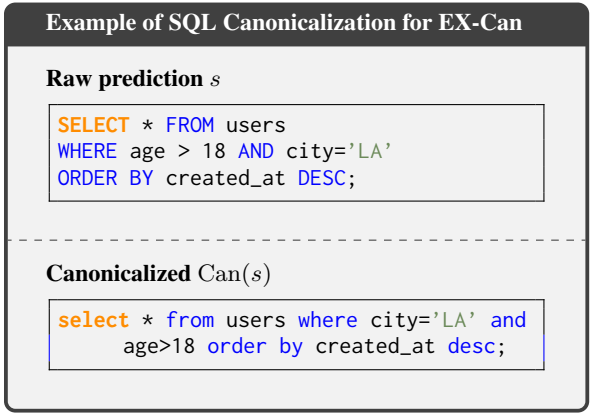

\centering
\small
\begin{tcolorbox}[
  fonttitle=\bfseries,
  title=Example of SQL Canonicalization for EX-Can,
  size=normal,
  % colback=black!4,
  % colframe=black,
  label=canonbox
]

\textbf{Raw prediction $s$}
\begin{lstlisting}
SELECT * FROM users
WHERE age > 18  AND  city='LA'
ORDER BY created_at DESC;
\end{lstlisting}

\tcbline

\textbf{Canonicalized $\mathrm{Can}(s)$}
\begin{lstlisting}
select * from users where city='LA' and age>18 order by created_at desc;
\end{lstlisting}

\end{tcolorbox}

\caption{An example of SQL canonicalization used for EX-Can. The procedure normalizes keyword case, whitespace, punctuation spacing, and reorders flat AND conditions within the same logical level.}
\label{fig:canon-example}
\end{figure}

\paragraph{Canonical Exact Match (EX-Can).}
We additionally report canonical exact match, which compares SQL strings after conservative normalization:
\begin{equation}
\text{EX-Can}=\frac{1}{N}\sum_{i=1}^{N}\mathbf{1}\!\left[\text{Can}(s_i)=\text{Can}(s_i^\ast)\right].
\end{equation}
Canonicalization removes surface-form differences while avoiding semantic changes. Details are provided in Appendix~\ref{alg:canonicalize}, with an illustrative example in Figure~\ref{fig:canon-example}.

%====================

\paragraph{Robustness Metrics.}
Robustness is measured as the drop in execution accuracy under visual perturbation $\mathcal{P}$:
\begin{equation}
\Delta_{\mathcal{P}}=\text{EXAcc}_{\text{clean}}-\text{EXAcc}_{\mathcal{P}}.
\end{equation}

\subsection{Experimental Conditions and Ablations}
\label{sec:exp-conditions}

We design a set of experiments to isolate representation sufficiency, visual grounding, and efficiency trade-offs.

\paragraph{Token budget sweep.}
We vary the optical token budget (e.g., 64, 100, 256, 400) while keeping all other components fixed, enabling systematic analysis of accuracy-efficiency trade-offs.

\paragraph{\textsc{FrozenEnc} vs.\ \textsc{FullFT}.}
We compare the default \textsc{FrozenEnc} setting against a \textsc{FullFT} (where the encoder is also updated) ablation in which both encoder and decoder are fine-tuned. \textsc{FullFT} is used only to study accuracy and robustness trade-offs and is not the default configuration.

\paragraph{No-image and wrong-image diagnostics.}
To verify visual grounding, we evaluate \textsc{NoImage} (optical tokens removed) and \textsc{WrongTable} (table images permuted across examples), testing whether models collapse into language-only or template-based generation.

\begin{figure*}[ht]
    \centering
    \includegraphics[width=\linewidth]{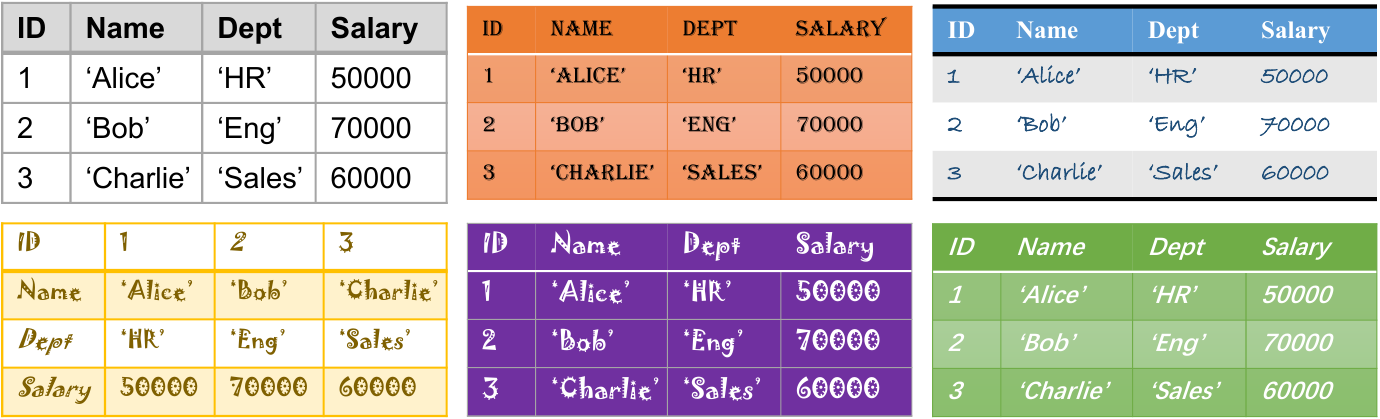}
    \caption{
    % Different styles of the same content of a table.
    Different visual renderings of the same table content. Variations include changes in font style, color, layout, borders, and spacing, while preserving identical schema and cell values. These styles are used to study robustness to superficial visual changes.
    }
    \label{fig:styles}
\end{figure*}

\subsection{Robustness Augmentation and Evaluation}
\label{sec:robustness-aug}

We study robustness under both training-time augmentation and inference-time perturbation. The two are intentionally separated to distinguish robustness gained through encoder adaptation from robustness intrinsic to the optical token interface.

\paragraph{Training-time augmentation (\textsc{FullFT}).}
When fine-tuning the visual encoder (\textsc{FullFT}), we apply rendering-based augmentations that preserve table semantics while altering visual appearance. These augmentations are designed to encourage invariance to superficial layout and style variations, and are applied on-the-fly during training.

Specifically, we use:
(i) \textit{Style variation}, including changes in font family and size, spacing, border thickness, zebra striping, cell padding, and column width; and
(ii) \textit{Table transposition}, which swaps rows and columns while relocating headers accordingly.
Examples of such augmentations are shown in Figure~\ref{fig:styles}.
The supervision signal (ground-truth SQL) remains unchanged.
Unless otherwise specified, these augmentations are enabled only in the \textsc{FullFT} ablation and are not used in the default \textsc{FrozenEnc} setting.

\paragraph{Inference-time robustness evaluation.}
To evaluate robustness independently of retraining, we apply controlled test-time perturbations to the rendered table images and measure the resulting drop in execution accuracy $\Delta_{\mathcal{P}}\!=\!\text{EXAcc}_{\text{clean}}\!-\!\text{EXAcc}_{\mathcal{P}}$.

All perturbations preserve table content except where explicitly noted.

\paragraph{StyleShift.}
\textit{StyleShift} samples a different rendering template at test time while keeping the underlying table content fixed. This perturbation assesses invariance to superficial visual factors such as font choice, spacing, and grid styling, without altering textual or relational information.

\paragraph{HeaderMask.}
\textit{HeaderMask} partially occludes column headers to disrupt explicit schema cues while preserving table body cells.
We mask only the header row region using opaque rectangular fills.
The masking ratio is capped at at most one-third of the header width per table.
Mask locations are sampled uniformly over contiguous header spans (or multiple short spans).
Empirically, larger masking ratios lead to unstable behavior and can prevent convergence, and are therefore excluded.

\paragraph{Visual grounding diagnostics.}
In addition to robustness perturbations, we include two diagnostic settings to test visual grounding.
\textsc{NoImage} removes optical tokens (or replaces them with a learned null token) while keeping the question unchanged.
\textsc{WrongTable} permutes table images across examples, producing mismatched table-question pairs.
These settings are not intended as realistic noise models, but as stress tests to verify that SQL generation does not collapse into language-only or template-based behavior.

Default parameterizations for all perturbations are summarized in Appendix~\ref{app:robust}.

\subsection{Implementation Details}
\label{sec:impl}

OptiSQL uses a pretrained OCR-oriented visual encoder to produce compact optical tokens. Unless otherwise specified, the encoder is frozen and only the autoregressive decoder is fine-tuned using token-level cross-entropy loss over SQL targets. For \textsc{FullFT}, both encoder and decoder are updated and robustness-oriented training-time augmentations are enabled. All models are evaluated under identical decoding and runtime settings.

% ====================================

\begin{table*}[t]
    \centering
    \resizebox{\linewidth}{!}{
    \begin{tabular}{l l cccccc}
    \toprule
    Model & Input Table & Tokens & EXAcc & EX-Can & EXAcc St. & EXAcc HM
    \\
    \midrule\midrule
    \rfc~\cite{reforce}
    & Text
    & 3,500
    & \textbf{0.78}
    & \textbf{0.86}
    & \textbf{0.86}
    & \textbf{0.86}
    \\
    \dso~\cite{deepseekocr}
    & OCR text
    & 3,500
    & 0.62
    & 0.70
    & 0.58
    & 0.45
    \\
    \pix~\cite{p2s}
    & Image
    & 800
    & 0.55
    & 0.63
    & 0.50
    & 0.38
    \\
    \don~\cite{donut}
    & Image
    & 900
    & 0.57
    & 0.65
    & 0.48
    & 0.35
    \\
    \textbf{\soe}
    & Image
    & \textbf{256}
    & 0.66
    & 0.76
    & \underline{0.72}
    & \underline{0.60}
    \\
    \textbf{\sof}
    & Image
    & \textbf{256}
    & \underline{0.68}
    & \underline{0.77}
    & 0.65
    & 0.55
    \\
    \bottomrule
    \end{tabular}
    }
    \caption{    Table Tokens count only tokens used to represent table information, excluding question and output tokens. ReFoRCE is a text-only upper-bound baseline. EXAcc St. represents the style-shift setting, and EXAcc HM denotes the header-masking setting. Best results are in \textbf{bold}, second-best are \underline{underlined}.}
    \label{tab:main}
\end{table*}

\section{Results and Analysis}
\label{sec:results}
We analyze OptiSQL with a focus on representation sufficiency rather than leaderboard performance. Specifically, we ask whether compact optical tokens can preserve sufficient structural and semantic information to support the generation of executable SQL under explicit efficiency constraints. Unless otherwise specified, OptiSQL refers to the default \textsc{FrozenEnc} setting.

\subsection{Main Results}
\label{res:main}
Table~\ref{tab:main} reports execution accuracy, canonical exact match, robustness, and table-related token cost across representative baselines. For all methods, we report only tokens used to encode table information, excluding question and output tokens.

Text-only systems achieve the highest execution accuracy when ground-truth textual schemas and table contents are available. In particular, ReFoRCE benefits from lossless access to explicit schema structure and therefore serves as an \textbf{upper-bound reference} rather than a directly comparable system. This setting assumes strictly stronger input supervision than any image-based formulation.

In contrast, OCR-based pipelines exhibit a substantial performance drop. Despite reconstructing text explicitly, OCR errors and schema linearization introduce compounding noise and long input sequences. General-purpose VLMs further underperform on executable accuracy, suggesting that document-level generation objectives do not directly translate to structured program synthesis.

Under a strictly weaker and more realistic assumption, where table information is provided only as images and accessed through compact optical tokens, OptiSQL achieves the strongest execution accuracy among all non-text-only approaches. With only 256 optical tokens, OptiSQL approaches the performance of OCR pipelines while using an order of magnitude fewer table tokens. These results demonstrate that optical tokens produced by an OCR-oriented encoder preserve sufficient structural and semantic information to support executable SQL generation under severe compression.

\paragraph{Robustness and encoder adaptation.}
Both \textsc{FrozenEnc} and \textsc{FullFT} degrade under test-time perturbations, with header masking consistently more damaging than style variation. While \textsc{FullFT} slightly improves clean accuracy, it exhibits larger robustness drops, indicating overfitting to surface rendering patterns even with training-time augmentation. This trade-off motivates our choice of \textsc{FrozenEnc} as the default setting for studying representation sufficiency and efficiency.

\subsection{Visual Grounding Diagnostics}
\label{res:ablation}
To verify that OptiSQL relies on visual table information rather than language priors, we conduct strong diagnostic ablations that remove or corrupt the visual input. Table~\ref{tab:grounding} reports execution accuracy for \textsc{FrozenEnc} under these conditions.

\begin{table}[t]
    \centering
    \resizebox{\linewidth}{!}{
    \begin{tabular}{lll}
    \toprule
    Setting & EXAcc (\%) & EX-Can (\%)
    \\
    \midrule\midrule
    Clean Image & 66 & 76
    \\
    NoImage & 15~\small{(\footnotesize{\textcolor{red}{-83.33\%}})} & 28~\small{(\footnotesize{\textcolor{red}{-63.16\%}})}
    \\
    RandomImage & 22~\small{(\footnotesize{\textcolor{red}{-66.67\%}})} & 30~\small{(\footnotesize{\textcolor{red}{-60.53\%}})}
    \\
    WrongTable & 6~\small{(\footnotesize{\textcolor{red}{-90.91\%}})} & 12~\small{(\footnotesize{\textcolor{red}{-84.21\%}})}
    \\
    \bottomrule
    \end{tabular}
    }
    \caption{
    EXAcc and EX-Can under visual grounding diagnostics. Removing, randomizing, or mismatching table images causes substantial performance degradation, highlighting the importance of correct visual grounding.
    }
    \label{tab:grounding}
\end{table}

\begin{figure}[t]
    \centering
    \includegraphics[width=\linewidth]{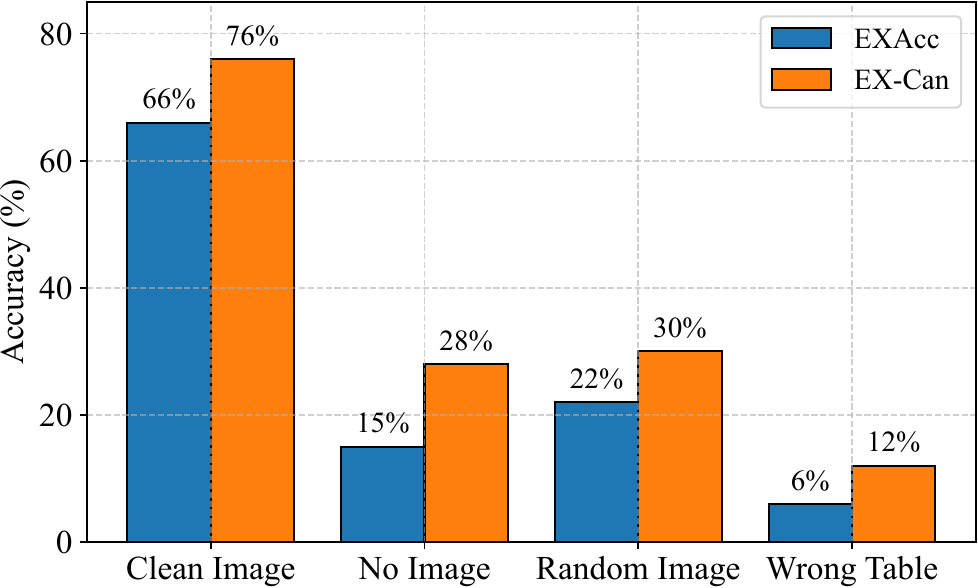}
    \caption{Visual grounding diagnostics for OptiSQL (FrozenEnc). EXAcc and EX-Can under clean and perturbed visual inputs. Bars show accuracy. Performance degrades sharply when visual grounding is disrupted.}
    \label{fig:ablation}
\end{figure}

Removing optical tokens (\textsc{NoImage}) causes a dramatic collapse in execution accuracy, confirming that SQL generation does not reduce to a language-only or template-based process. Replacing the table image with random or mismatched images further degrades performance, demonstrating that OptiSQL is sensitive to table-specific visual content rather than generic visual cues. Together, these results provide direct evidence that optical tokens act as a necessary conditioning signal for executable reasoning.

\subsection{Efficiency-Accuracy Trade-offs}
\label{res:token}
\begin{figure}[t]
    \centering
    \includegraphics[width=\linewidth]{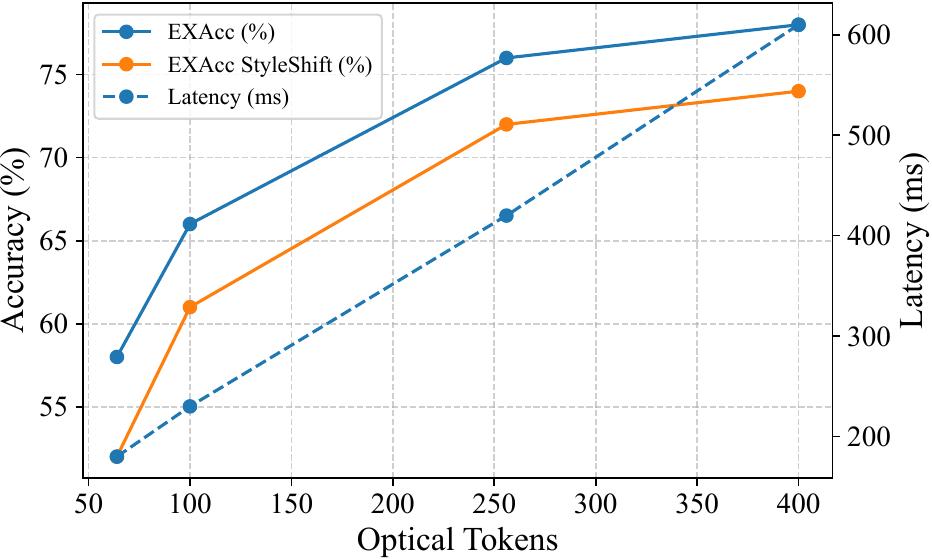}
    \caption{
    Execution accuracy and inference latency as a function of the optical token budget. Increasing the number of optical tokens improves execution accuracy with diminishing returns beyond 256 tokens, while latency grows approximately linearly, revealing a clear efficiency-accuracy trade-off.
    }
    \label{fig:tokens}
\end{figure}

We next analyze how performance scales with the optical token budget. Figure~\ref{fig:tokens} shows execution accuracy and latency across budgets.

Execution accuracy increases monotonically as the token budget grows, with diminishing returns beyond approximately 256 tokens. This identifies a clear efficiency sweet spot where OptiSQL achieves strong executable accuracy at moderate runtime cost. Notably, robustness trends remain stable across budgets, indicating that compression primarily limits representational capacity rather than altering sensitivity to visual perturbations.

\paragraph{Scope of Comparison.}
OptiSQL is not designed to compete directly with leaderboard systems on Spider~2.0, which assume full access to ground-truth textual schemas and contents. Such comparisons would conflate differences in task formulation with performance in modeling. Instead, we include a strong text-only system as an upper-bound reference to contextualize the efficiency-accuracy trade-off introduced by optical representations. Our objective is to study whether executable reasoning remains feasible when table information is accessed through a highly compressed visual channel.

\subsection{Summary}
\label{res:disc}

Overall, these results show that compact optical tokens can serve as an effective and analyzable interface for executable semantic parsing. Compared to textual representations, they offer substantial compression at a controlled accuracy cost. Compared to OCR pipelines, they avoid explicit reconstruction and reduce error propagation. Compared to general-purpose vision-language models, OptiSQL benefits from OCR-oriented tokenization and task-specific decoder adaptation. Rather than replacing text-based settings when ground-truth schemas are available, optical tokens enable a principled exploration of efficiency-oriented semantic parsing under realistic input constraints.

\section{Conclusion}
\label{sec:conclusion}

We study executable SQL generation under a realistic setting where tables are available only as visual artifacts rather than structured text. Instead of reconstructing tables into long textual inputs, we investigate whether compact optical tokens can serve as an efficient interface for semantic parsing.
Experiments on a visualized version of Spider~2.0-Snow show that optical tokens achieve substantial input compression while preserving strong execution accuracy. Compared to OCR-based pipelines and general-purpose vision-language models, OptiSQL offers a more favorable efficiency-accuracy trade-off under identical input assumptions.
These results suggest that optical tokenization is a practical alternative to text-centric table representations when structured schemas are unavailable, and highlight representation choice as a key design for executable reasoning under long-context constraints.

\section*{Limitations}
\label{sec:limit}
OptiSQL relies on a pretrained visual encoder, and its performance is bounded by the encoder's ability to represent textual and structural information in table images. Inputs with severe noise, handwriting, or distortions may not be well handled.
Moreover, our evaluation focuses on single-table queries in a visualized Spider~2.0-Snow setting. Extending the approach to multi-table queries will require modeling inter-table relations and joins, which we leave to future work.
Finally, we primarily adopt a frozen-encoder training protocol to study representation sufficiency. While fine-tuning the encoder can improve accuracy, it may reduce robustness, suggesting a trade-off that warrants investigation.
We focus exclusively on executable SQL generation; the applicability of compact optical tokens to other structured generation tasks remains open.

% \input{sections/X02_Acknowledgements}

\bibliography{custom}

@misc{reforce,
      title={{ReFoRCE}: A Text-to-SQL Agent with Self-Refinement, Consensus Enforcement, and Column Exploration}, 
      author={Minghang Deng and Ashwin Ramachandran and Canwen Xu and Lanxiang Hu and Zhewei Yao and Anupam Datta and Hao Zhang},
      year={2025},
      eprint={2502.00675},
      archivePrefix={arXiv},
      primaryClass={cs.CL},
      url={https://arxiv.org/abs/2502.00675}, 
}

@misc{deepseekocr,
      title={DeepSeek-OCR: Contexts Optical Compression}, 
      author={Haoran Wei and Yaofeng Sun and Yukun Li},
      year={2025},
      eprint={2510.18234},
      archivePrefix={arXiv},
      primaryClass={cs.CV},
      url={https://arxiv.org/abs/2510.18234}, 
}

@misc{p2s,
      title={Pix2Struct: Screenshot Parsing as Pretraining for Visual Language Understanding}, 
      author={Kenton Lee and Mandar Joshi and Iulia Turc and Hexiang Hu and Fangyu Liu and Julian Eisenschlos and Urvashi Khandelwal and Peter Shaw and Ming-Wei Chang and Kristina Toutanova},
      year={2023},
      eprint={2210.03347},
      archivePrefix={arXiv},
      primaryClass={cs.CL},
      url={https://arxiv.org/abs/2210.03347}, 
}

@inproceedings{donut,
  title     = {OCR-Free Document Understanding Transformer},
  author    = {Kim, Geewook and Hong, Teakgyu and Yim, Moonbin and Nam, JeongYeon and Park, Jinyoung and Yim, Jinyeong and Hwang, Wonseok and Yun, Sangdoo and Han, Dongyoon and Park, Seunghyun},
  booktitle = {European Conference on Computer Vision (ECCV)},
  year      = {2022}
}

@inproceedings{yu2018spider,
    title = "{S}pider: A Large-Scale Human-Labeled Dataset for Complex and Cross-Domain Semantic Parsing and Text-to-{SQL} Task",
    author = "Yu, Tao  and
      Zhang, Rui  and
      Yang, Kai  and
      Yasunaga, Michihiro  and
      Wang, Dongxu  and
      Li, Zifan  and
      Ma, James  and
      Li, Irene  and
      Yao, Qingning  and
      Roman, Shanelle  and
      Zhang, Zilin  and
      Radev, Dragomir",
    editor = "Riloff, Ellen  and
      Chiang, David  and
      Hockenmaier, Julia  and
      Tsujii, Jun{'}ichi",
    booktitle = "Proceedings of the 2018 Conference on Empirical Methods in Natural Language Processing",
    month = oct # "-" # nov,
    year = "2018",
    address = "Brussels, Belgium",
    publisher = "Association for Computational Linguistics",
    url = "https://aclanthology.org/D18-1425/",
    doi = "10.18653/v1/D18-1425",
    pages = "3911--3921"
}

@misc{spider2_arxiv,
  title        = {Spider 2.0: Evaluating Language Models on Real-World Enterprise Text-to-{SQL} Workflows},
author={Fangyu Lei and Jixuan Chen and Yuxiao Ye and Ruisheng Cao and Dongchan Shin and Hongjin Su and Zhaoqing Suo and Hongcheng Gao and Wenjing Hu and Pengcheng Yin and Victor Zhong and Caiming Xiong and Ruoxi Sun and Qian Liu and Sida Wang and Tao Yu},
      year={2025},
      eprint={2411.07763},
      archivePrefix={arXiv},
      primaryClass={cs.CL},
      url={https://arxiv.org/abs/2411.07763}, 
}

@article{liu2023lostmiddle,
    title = "Lost in the Middle: How Language Models Use Long Contexts",
    author = "Liu, Nelson F.  and
      Lin, Kevin  and
      Hewitt, John  and
      Paranjape, Ashwin  and
      Bevilacqua, Michele  and
      Petroni, Fabio  and
      Liang, Percy",
    journal = "Transactions of the Association for Computational Linguistics",
    volume = "12",
    year = "2024",
    address = "Cambridge, MA",
    publisher = "MIT Press",
    url = "https://aclanthology.org/2024.tacl-1.9/",
    doi = "10.1162/tacl_a_00638",
    pages = "157--173"
}

@misc{cai2025text2sqlflowrobustsqlawaredata,
      title={Text2SQL-Flow: A Robust SQL-Aware Data Augmentation Framework for Text-to-SQL}, 
      author={Qifeng Cai and Hao Liang and Chang Xu and Tao Xie and Wentao Zhang and Bin Cui},
      year={2025},
      eprint={2511.10192},
      archivePrefix={arXiv},
      primaryClass={cs.CL},
      url={https://arxiv.org/abs/2511.10192}, 
}

@misc{wang2025querylevelcomparisonfinegrainedreinforcement,
      title={Beyond Query-Level Comparison: Fine-Grained Reinforcement Learning for Text-to-SQL with Automated Interpretable Critiques}, 
      author={Guifeng Wang and Yuanfeng Song and Meng Yang and Tao Zhu and Xiaoming Yin and Xing Chen},
      year={2025},
      eprint={2511.22258},
      archivePrefix={arXiv},
      primaryClass={cs.CL},
      url={https://arxiv.org/abs/2511.22258}, 
}

@misc{guo2019complextexttosqlcrossdomaindatabase,
      title={Towards Complex Text-to-SQL in Cross-Domain Database with Intermediate Representation}, 
      author={Jiaqi Guo and Zecheng Zhan and Yan Gao and Yan Xiao and Jian-Guang Lou and Ting Liu and Dongmei Zhang},
      year={2019},
      eprint={1905.08205},
      archivePrefix={arXiv},
      primaryClass={cs.CL},
      url={https://arxiv.org/abs/1905.08205}, 
}

@inproceedings{li-etal-2020-tablebank,
    title = "{T}able{B}ank: Table Benchmark for Image-based Table Detection and Recognition",
    author = "Li, Minghao  and
      Cui, Lei  and
      Huang, Shaohan  and
      Wei, Furu  and
      Zhou, Ming  and
      Li, Zhoujun",
    editor = "Calzolari, Nicoletta  and
      B{\'e}chet, Fr{\'e}d{\'e}ric  and
      Blache, Philippe  and
      Choukri, Khalid  and
      Cieri, Christopher  and
      Declerck, Thierry  and
      Goggi, Sara  and
      Isahara, Hitoshi  and
      Maegaard, Bente  and
      Mariani, Joseph  and
      Mazo, H{\'e}l{\`e}ne  and
      Moreno, Asuncion  and
      Odijk, Jan  and
      Piperidis, Stelios",
    booktitle = "Proceedings of the Twelfth Language Resources and Evaluation Conference",
    month = may,
    year = "2020",
    address = "Marseille, France",
    publisher = "European Language Resources Association",
    url = "https://aclanthology.org/2020.lrec-1.236/",
    pages = "1918--1925",
    language = "eng",
    ISBN = "979-10-95546-34-4"
}

@misc{doc2table,
	author = {},
	title = {{D}oc2{T}able --- doc2table.com},
	howpublished = {\url{https://www.doc2table.com/home}},
	year = {},
	note = {[Accessed 06-01-2026]},
}

@misc{sheng2024pdftableunifiedtoolkitdeep,
      title={PdfTable: A Unified Toolkit for Deep Learning-Based Table Extraction}, 
      author={Lei Sheng and Shuai-Shuai Xu},
      year={2024},
      eprint={2409.05125},
      archivePrefix={arXiv},
      primaryClass={cs.CV},
      url={https://arxiv.org/abs/2409.05125}, 
}

@inproceedings{guan-etal-2025-prep,
    title = "{P}re{P}-{OCR}: A Complete Pipeline for Document Image Restoration and Enhanced {OCR} Accuracy",
    author = "Guan, Shuhao  and
      Lin, Moule  and
      Xu, Cheng  and
      Liu, Xinyi  and
      Zhao, Jinman  and
      Fan, Jiexin  and
      Xu, Qi  and
      Greene, Derek",
    editor = "Che, Wanxiang  and
      Nabende, Joyce  and
      Shutova, Ekaterina  and
      Pilehvar, Mohammad Taher",
    booktitle = "Proceedings of the 63rd Annual Meeting of the Association for Computational Linguistics (Volume 1: Long Papers)",
    month = jul,
    year = "2025",
    address = "Vienna, Austria",
    publisher = "Association for Computational Linguistics",
    url = "https://aclanthology.org/2025.acl-long.749/",
    doi = "10.18653/v1/2025.acl-long.749",
    pages = "15413--15425",
    ISBN = "979-8-89176-251-0"
}

@inproceedings{shi-etal-2025-gen,
    title = "Gen-{SQL}: Efficient Text-to-{SQL} By Bridging Natural Language Question And Database Schema With Pseudo-Schema",
    author = "Shi, Jie  and
      Xu, Bo  and
      Liang, Jiaqing  and
      Xiao, Yanghua  and
      Chen, Jia  and
      Xie, Chenhao  and
      Wang, Peng  and
      Wang, Wei",
    editor = "Rambow, Owen  and
      Wanner, Leo  and
      Apidianaki, Marianna  and
      Al-Khalifa, Hend  and
      Eugenio, Barbara Di  and
      Schockaert, Steven",
    booktitle = "Proceedings of the 31st International Conference on Computational Linguistics",
    month = jan,
    year = "2025",
    address = "Abu Dhabi, UAE",
    publisher = "Association for Computational Linguistics",
    url = "https://aclanthology.org/2025.coling-main.256/",
    pages = "3794--3807"
}

@misc{zhou2024dbgpthubopenbenchmarkingtexttosql,
      title={DB-GPT-Hub: Towards Open Benchmarking Text-to-SQL Empowered by Large Language Models}, 
      author={Fan Zhou and Siqiao Xue and Danrui Qi and Wenhui Shi and Wang Zhao and Ganglin Wei and Hongyang Zhang and Caigai Jiang and Gangwei Jiang and Zhixuan Chu and Faqiang Chen},
      year={2024},
      eprint={2406.11434},
      archivePrefix={arXiv},
      primaryClass={cs.DB},
      url={https://arxiv.org/abs/2406.11434}, 
}

\appendix
\clearpage
\section{Appendix}
\label{sec:appendix}

This appendix provides implementation and protocol details that are omitted from the main text for conciseness. In particular, we describe (1) token accounting and TSR computation (table-only), (2) dataset visualization and rendering pipeline following DeepSeek-OCR style inputs, (3) robustness perturbation generation, (4) execution evaluation protocol in SQLite with failure taxonomy, (5) training and decoding hyperparameters for \textsc{FrozenEnc} and \textsc{FullFT}, and (6) additional canonicalization examples and intentionally excluded rules.

% =========================================================
\subsection{Canonicalization for EX-Can}
\label{app:canon}

We compute Canonical Exact Match (EX-Can) by applying a conservative canonicalization function $\text{Can}(\cdot)$ to both predicted and gold SQL strings, and then performing exact string comparison. Canonicalization removes surface-form differences that are unlikely to change semantics, while avoiding aggressive rewrites that may introduce incorrect equivalence.

\begin{algorithm}[t]
\small
\caption{Canonicalization pipeline for EX-Can}
\label{alg:canonicalize}
\KwIn{SQL string $s$}
\KwOut{Canonicalized SQL string $\text{Can}(s)$}

\BlankLine
\textbf{Optional preprocessing:}\\
$s \leftarrow \textsc{StripComments}(s)$ \tcp*{remove SQL comments if present}
$s \leftarrow \textsc{NormalizeUnicode}(s)$ \tcp*{normalize Unicode characters}

\BlankLine
\textbf{Step 1 (Case normalization):}\\
$s \leftarrow \textsc{LowercaseKeywords}(s)$ \tcp*{or uppercase, applied consistently}

\BlankLine
\textbf{Step 2 (Whitespace normalization):}\\
$s \leftarrow \textsc{CollapseWhitespace}(s)$ \tcp*{spaces/newlines/tabs $\rightarrow$ single space}
$s \leftarrow \textsc{Trim}(s)$

\BlankLine
\textbf{Step 3 (Punctuation spacing):}\\
$s \leftarrow \textsc{NormalizeOpSpacing}(s)$ \tcp*{e.g., \texttt{a=1} $\leftrightarrow$ \texttt{a = 1}}

\BlankLine
\textbf{Step 4 (Redundant parentheses):}\\
$s \leftarrow \textsc{RemoveRedundantParens}(s)$ \tcp*{only if precedence is preserved}

\BlankLine
\textbf{Step 5 (AND/OR condition ordering):}
\begin{enumerate}\small
\item Parse $s$ into an abstract syntax tree (AST): $T \leftarrow \textsc{ParseSQL}(s)$
\item For each \texttt{WHERE} clause node in $T$:
  \begin{enumerate}\small
  \item Identify maximal \emph{flat} boolean chains at the same syntactic level (handle \texttt{AND} and \texttt{OR} separately)
  \item Canonicalize each condition in the chain to a string key
  \item Sort conditions lexicographically within each chain
  \item Reconstruct the chain without re-associating nested boolean expressions
  \end{enumerate}
\item Serialize the modified AST back to SQL: $s \leftarrow \textsc{SerializeSQL}(T)$
\end{enumerate}

\BlankLine
\Return{$s$}
\end{algorithm}

\paragraph{More examples.}
Below are common cases handled by our pipeline:
\begin{itemize}
  \item \textbf{Whitespace and case:}
  \texttt{SELECT  *  FROM T} $\rightarrow$ \texttt{select * from t}.
  \item \textbf{Operator spacing:}
  \texttt{a=1 AND b =2} $\rightarrow$ \texttt{a=1 and b=2}.
  \item \textbf{Flat boolean ordering (same level only):}
  \texttt{where age>18 and city='LA'} $\leftrightarrow$
  \texttt{where city='LA' and age>18}.
\end{itemize}

\paragraph{Rules we intentionally do \emph{not} apply.}
We do not reorder JOINs, do not normalize nested boolean structures (beyond flat chains), do not rewrite subqueries, and do not apply algebraic transformations such as distributing predicates across OR clauses. These operations can change semantics or require database-aware reasoning, and are therefore excluded.

% =========================================================
\subsection{Token Accounting and TSR (Table-only)}
\label{app:token}

In the main text we report input efficiency with table-only token accounting. That is, we measure how many tokens are needed to represent \emph{only the table}, and we ignore question tokens and output tokens for TSR and related statistics. This choice isolates the compression benefit of optical tokenization on tabular content.

\paragraph{Table-text token length.}
Let $\text{Lin}(\cdot)$ denote a deterministic textual linearization of the table (schema plus cell values) used by the text-only baseline and by the OCR pipeline after reconstruction. We compute
\begin{equation}
  L_{\text{table-text}}(i) = \left|\text{Tok}\!\left(\text{Lin}(\text{table}_i)\right)\right|,
\end{equation}
where $\text{Tok}(\cdot)$ is the tokenizer of the decoder (same tokenizer used for SQL generation), and $|\cdot|$ counts tokens.

\paragraph{Optical token length.}
For OptiSQL, let $V_i = E(I_i)$ be the optical token sequence produced by the OCR-oriented encoder. We compute
\begin{equation}
  L_{\text{opt}}(i) = |V_i|.
\end{equation}
Under a fixed token budget configuration (for example 256), $L_{\text{opt}}(i)$ is bounded and typically constant for most samples, except when dynamic tiling is enabled.

\paragraph{Token Savings Ratio (TSR).}
We define TSR strictly on the table representation:
\begin{equation}
  \text{TSR} =
  \frac{\mathbb{E}_i\left[L_{\text{table-text}}(i)\right]}
       {\mathbb{E}_i\left[L_{\text{opt}}(i)\right]}.
\end{equation}
For interpretability, we also report the per-query averages $\mathbb{E}[L_{\text{table-text}}]$ and $\mathbb{E}[L_{\text{opt}}]$.

\paragraph{Implementation notes.}
\begin{itemize}
  \item We use the \emph{same} tokenizer across baselines to avoid tokenizer-induced artifacts.
  \item $\text{Lin}(\cdot)$ is deterministic, including fixed column ordering and row ordering, so that TSR is stable across runs.
  \item For OCR pipelines, $L_{\text{table-text}}$ is computed on the reconstructed text fed into the downstream text-to-SQL model. This number is therefore comparable to the text-only baseline.
\end{itemize}

% =========================================================
\subsection{Visualized Spider 2.0-Snow Construction and Rendering}
\label{app:render}

We build a visualized version of Spider 2.0-Snow~\cite{spider2_arxiv} by rendering database tables into images while keeping natural language questions and gold SQL unchanged. Rendering follows the image-input conventions used by DeepSeek-OCR~\cite{deepseekocr} to ensure compatibility with OCR-oriented optical tokenization.

\paragraph{Table extraction.}
For each example, we identify the target table(s) required by the gold SQL. In our single-table setting, we extract the referenced table into a structured grid with:
\begin{itemize}
  \item a header row (column names),
  \item body cells (stringified values with type-aware formatting),
  \item optional row index column (disabled by default, enabled only for transpose augmentation).
\end{itemize}

\paragraph{Base rendering pipeline.}
We render tables via an HTML-CSS template and export to PNG:
\begin{enumerate}
  \item Serialize the table grid into HTML.
  \item Apply a CSS theme (fonts, border rules, padding, alignment).
  \item Use a headless browser to rasterize the HTML into a high-resolution PNG with a white background and tight crop.
  \item Apply model-side preprocessing expected by the encoder (RGB conversion, resize with aspect ratio preserved, and padding if needed).
\end{enumerate}
Unless otherwise specified, \textsc{FrozenEnc} uses a single base style at both training and evaluation.

\paragraph{Style templates (for \textsc{FullFT} augmentation).}
For training-time style augmentation (\textsc{FullFT} only), we sample from a set of style templates that vary:
\begin{itemize}
  \item font family and size, boldness,
  \item cell padding, row height, and column width,
  \item border thickness and grid visibility,
  \item zebra striping and header emphasis.
\end{itemize}
Figure~\ref{fig:styles} shows representative examples in the main text.

\paragraph{Table transpose augmentation (\textsc{FullFT} only).}
We optionally apply a transpose transform to encourage invariance to layout changes while preserving table semantics. Concretely, we render a transposed view in which each original column becomes a labeled row, and we add an index header for original rows. This produces a visually distinct but information-preserving representation. The supervision remains the same gold SQL because the underlying data content and header-value associations are preserved in the rendered view.

% =========================================================
\subsection{Robustness Perturbations}
\label{app:robust}

We evaluate robustness using test-time perturbations that modify the rendered table image without retraining. All perturbations preserve table content except for explicit occlusion in \textit{HeaderMask}.

\paragraph{StyleShift.}
\textit{StyleShift} samples a different rendering template at test time while preserving the same grid content. This assesses invariance to superficial style factors.

\paragraph{HeaderMask.}
\textit{HeaderMask} occludes a portion of column headers. We mask only the header row region (not body cells) with a solid rectangle fill. The masking ratio is capped at $1/3$ of the header region. Empirically, larger masking ratios cause optimization instability and can prevent convergence.

\begin{itemize}
  \item \textbf{Mask target:} header row cells only.
  \item \textbf{Mask ratio cap:} at most $1/3$ of header width per table.
  \item \textbf{Sampling:} uniformly sample a contiguous span (or multiple short spans) over header cells, then draw opaque rectangles.
\end{itemize}

\paragraph{NoImage and WrongTable.}
\textsc{NoImage} removes optical tokens (or replaces them with a learned null token) while keeping the question unchanged. \textsc{WrongTable} permutes table images across examples, producing mismatched table-question pairs. Both are diagnostic tests for visual grounding.

\paragraph{Suggested perturbation parameterization.}
Table~\ref{tab:perturb-params} summarizes default parameters used in our experiments.

\begin{table}[t]
\centering
\small
\resizebox{\linewidth}{!}{
\begin{tabular}{l l}
\toprule
Perturbation & Default parameters \\
\midrule
StyleShift & sample 1 template uniformly from style pool \\
HeaderMask & mask ratio $r \sim \text{Unif}(0, 1/3)$; header row only \\
NoImage & remove optical tokens, keep question tokens \\
WrongTable & permute table images within evaluation batch or dataset \\
\bottomrule
\end{tabular}
}
\caption{Default robustness perturbation parameters used in evaluation.}
\label{tab:perturb-params}
\end{table}

% =========================================================
\subsection{Execution Evaluation Protocol in SQLite}
\label{app:exec}

We evaluate executable correctness by executing both predicted SQL and gold SQL in SQLite, using the same database instance per example. Execution timeouts are treated as failures.

\paragraph{Engine and safety.}
We use SQLite as the DB engine. For each query:
\begin{enumerate}
  \item open a read-only connection to the corresponding database,
  \item execute the gold SQL to obtain the reference result $R^\ast$,
  \item execute the predicted SQL to obtain $R$ if execution succeeds.
\end{enumerate}

\paragraph{Timeout handling.}
We enforce a wall-clock timeout (for example 2 seconds) per query using a process-level timeout mechanism. If a query exceeds the timeout, it is marked as \textit{Timeout} and counted as execution failure. After excluding device instability, we treat timeouts as model-induced failures, typically caused by missing predicates or pathological joins.

\paragraph{Result normalization.}
We compare results after normalization:
\begin{itemize}
  \item treat row order as irrelevant unless an explicit \texttt{ORDER BY} exists in both queries,
  \item canonicalize numeric formatting (float tolerance if needed),
  \item normalize NULL and string encodings.
\end{itemize}

\paragraph{Failure taxonomy and additional statistics.}
Beyond EXAcc, we compute:
\begin{itemize}
  \item \textbf{ValidSQL}: percentage of predicted SQL that executes successfully within timeout,
  \item \textbf{NonExecutable}: percentage with SQL syntax errors, missing tables or columns, or other runtime errors,
  \item \textbf{Timeout}: percentage exceeding the timeout.
\end{itemize}

We observed that among the failed cases (EXAcc $= 0$), non-executable predictions account for roughly 50\%. Table~\ref{tab:exec-breakdown} reports a representative breakdown consistent with this observation. These values should be updated with the final run logs, but the taxonomy and reporting format remain the same.

\begin{table}[t]
\centering
\small
\resizebox{\linewidth}{!}{
\begin{tabular}{l c}
\toprule
Outcome category & Rate (\%) \\
\midrule
Correct execution (EXAcc = 1) & 76 \\
Executable but wrong result & 10 \\
Timeout & 2 \\
Non-executable (syntax or schema error) & 12 \\
\midrule
Total failures (EXAcc = 0) & 24 \\
Non-executable among failures & 50 \\
\bottomrule
\end{tabular}
}
\caption{Execution outcome breakdown under SQLite. Among failures, non-executable predictions are about half.}
\label{tab:exec-breakdown}
\end{table}

\paragraph{Common failure causes.}
We find two dominant sources:
\begin{itemize}
  \item \textbf{Non-executable:} schema linking errors (wrong column names, missing table aliases), malformed quoting, and invalid aggregation usage.
  \item \textbf{Executable but wrong:} incorrect join paths, missing filter predicates, and wrong aggregation or grouping.
\end{itemize}
Timeouts are commonly triggered by missing WHERE predicates leading to full scans, or unintended Cartesian products.

% =========================================================
\subsection{Training and Decoding Hyperparameters (\textsc{FrozenEnc} vs \textsc{FullFT})}
\label{app:hyper}

This section provides a complete hyperparameter specification used for the two training variants:
\textsc{FrozenEnc} (default) updates only the decoder, and \textsc{FullFT} updates both encoder and decoder with training-time augmentations.

\paragraph{Optimizer and schedule.}
We use AdamW for all experiments. We recommend cosine decay with linear warmup. Gradient clipping is applied for stability.

\begin{table}[t]
\centering
\small
\begin{tabular}{l c c}
\toprule
Hyperparameter & \textsc{FrozenEnc} & \textsc{FullFT} \\
\midrule
Optimizer & AdamW & AdamW \\
$\beta_1, \beta_2$ & (0.9, 0.95) & (0.9, 0.95) \\
Weight decay & 0.1 & 0.1 \\
Warmup ratio & 0.03 & 0.05 \\
LR schedule & cosine decay & cosine decay \\
Grad clip norm & 1.0 & 1.0 \\
Precision & bf16 (or fp16) & bf16 (or fp16) \\
\bottomrule
\end{tabular}
\caption{Common optimization settings.}
\label{tab:opt-common}
\end{table}

\paragraph{Learning rates.}
\textsc{FrozenEnc} uses a higher decoder LR since only the decoder is trained. \textsc{FullFT} uses a much smaller encoder LR to reduce overfitting to visual styles and to preserve the pretrained optical tokenization behavior.

\begin{table}[t]
\centering
\small
\resizebox{\linewidth}{!}{
\begin{tabular}{l c c}
\toprule
Learning rate & \textsc{FrozenEnc} & \textsc{FullFT} \\
\midrule
Decoder LR & $2\times10^{-4}$ & $1\times10^{-4}$ \\
Encoder LR & 0 (frozen) & $5\times10^{-6}$ \\
Encoder LR relative scale & 0 & 20$\times$ smaller than decoder \\
\bottomrule
\end{tabular}
}
\caption{Recommended learning rates for \textsc{FrozenEnc} and \textsc{FullFT}.}
\label{tab:lr}
\end{table}

\paragraph{Batching and steps.}
We recommend setting global batch size via gradient accumulation to fit GPU memory. A practical configuration is:
\begin{itemize}
  \item \textbf{Global batch size:} 256 examples (via accumulation).
  \item \textbf{Training steps:} 20k to 30k for \textsc{FrozenEnc}, 10k to 20k for \textsc{FullFT} (higher cost per step).
\end{itemize}

\paragraph{Sequence lengths.}
We cap input and output lengths to stabilize training and ensure comparable latency reporting.
\begin{itemize}
  \item \textbf{Optical tokens:} $\{64, 100, 256, 400\}$ depending on budget.
  \item \textbf{Question length cap:} 128 tokens.
  \item \textbf{Max input length:} 512 tokens (optical plus question plus special tokens).
  \item \textbf{Max output length:} 256 tokens for SQL generation.
\end{itemize}

\paragraph{Decoding strategy.}
We use deterministic decoding for reliable executable evaluation.
\begin{itemize}
  \item \textbf{Default:} beam search with beam size 4, temperature 0, max output length 256.
  \item \textbf{Latency reporting:} greedy decoding (temperature 0) with the same max output length.
\end{itemize}

\paragraph{Augmentations in \textsc{FullFT}.}
Training-time augmentations described in Appendix~\ref{app:render} and Appendix~\ref{app:robust} (style variation and transpose) are enabled only for \textsc{FullFT}. All other settings are kept identical to \textsc{FrozenEnc} to isolate the effect of encoder adaptation.

% =========================================================
\subsection{Notes on Baseline Token Accounting}
\label{app:baseline-token}

To avoid confusion in Table 1, we emphasize that the OCR pipeline baseline consumes \emph{text} tokens after OCR reconstruction and schema linearization. Therefore, its Tokens/Query reflects the length of reconstructed table text (table-only) rather than optical tokens. In the main table, we recommend labeling this row as \textit{OCR-text + NL} and explicitly stating that its token count is computed under the same table-text token accounting defined in Appendix~\ref{app:token}.

\end{document}